\documentclass[11pt,a4paper]{article}
\usepackage[hyperref]{emnlp2020}
\usepackage{times}
\usepackage{latexsym}

\usepackage{graphicx}
\usepackage{subfigure}
\usepackage{booktabs} 
\usepackage{multirow}
\usepackage{adjustbox}
\usepackage{amsfonts,amssymb}
\usepackage{amsmath}

\usepackage{microtype}

\aclfinalcopy 
\def\aclpaperid{3273} 

\title{Beyond Language: Learning Commonsense from Images for Reasoning}

\author{Wanqing Cui, Yanyan Lan*, Liang Pang, Jiafeng Guo, Xueqi Cheng \\
CAS Key Lab of Network Data Science and Technology, \\ Institute of Computing Technology, Chinese Academy of Sciences, Beijing, China\\
University of Chinese Academy of Sciences, Beijing, China\\
\texttt{cuiwanqing18z, lanyanyan, pangliang, guojiafeng, cxq@ict.ac.cn}\\}

\date{}

\begin{document}
\maketitle
\begin{abstract}
This paper proposes a novel approach to learn commonsense from images, instead of limited raw texts or costly constructed knowledge bases, for the commonsense reasoning problem in NLP. Our motivation comes from the fact that an image is worth a thousand words, where richer scene information could be leveraged to help distill the commonsense knowledge, which is often hidden in languages. Our approach, namely Loire, consists of two stages. In the first stage, a bi-modal sequence-to-sequence approach is utilized to conduct the scene layout generation task, based on a text representation model ViBERT. In this way, the required visual scene knowledge, such as spatial relations, will be encoded in ViBERT by the supervised learning process with some bi-modal data like COCO. Then ViBERT is concatenated with a pre-trained language model to perform the downstream commonsense reasoning tasks. Experimental results on two commonsense reasoning problems, i.e.~commonsense question answering and pronoun resolution, demonstrate that Loire outperforms traditional language-based methods. We also give some case studies to show what knowledge is learned from images and explain how the generated scene layout helps the commonsense reasoning process.
\let\thefootnote\relax\footnotetext{*Corresponding Author}

\end{abstract}

\section{Introduction}

Commonsense reasoning is an important yet challenging task in artificial intelligence and natural language processing. Take commonsense question answering as an example, given a question and multiple choices, some commonsense knowledge is usually required to make the correct answer from the provided choices. Table~\ref{examples} show some typical commonsense question answering examples extracted from the dataset of commonsenseQA~\cite{talmor2018commonsenseqa}. 

\begin{table}
\caption{Examples from CommonsenseQA dataset.}
\label{examples}
\begin{center}
\begin{small}
\begin{tabular}{l p{5cm}}
\toprule
Question: & Where is a good idea but not required to have a fire extinguisher?\\
Choices: & (A)~school bus (B)~boat \textbf{(C)~house} (D)~hospital (E)~school\\
\midrule
Question: & Where can you put a picture frame when it's not hung vertically?\\
Choices: & (A)~art show (B)~wall (C)~newspaper (D)~car \textbf{(E)~table}\\
\bottomrule
\end{tabular}
\end{small}
\end{center}
\vskip -0.25in
\end{table}

Existing commonsense reasoning methods mainly utilize raw texts to conduct the data representation and answer prediction process~\cite{talmor2018commonsenseqa, rajani2019explain}. However, the background knowledge required in the commonsense reasoning task, such as spatial relations, causes and effects, scientific facts and social conventions, are usually not explicitly provided by the text. Therefore, it is difficult to capture such knowledge solely from the raw texts. Some other works propose to leverage knowledge bases to extract related commonsense knowledge~\cite{lin2019kagnet, lv2019graph, kipf2016semi, ye2019align, li2019teaching, ma2019towards}. However, the construction of a knowledge base is expensive, and the contained knowledge is too limited to fulfill the requirement. Furthermore, most commonsense question answering datasets, such as CommonsenseQA, are constructed from an existing knowledge base, e.g., ConceptNet \cite{speer2017conceptnet}. So it is unfair to use the knowledge base in these tasks. To sum up, how to automatically learn commonsense remains a challenging problem in NLP.

Motivated by the fact that images usually contain richer scene information, which can be viewed as an important supplementary resource to perceive for commonsense knowledge, this paper proposes to learn commonsense from images and incorporate such knowledge into the commonsense reasoning process. Take the question `\emph{Where is a good idea but not required to have a fire extinguisher?}' shown in Table~\ref{examples} as an example. Solving this problem requires a strong background knowledge that fire extinguishers are usually equipped in public places, such as hospitals, schools, and school buses. We can see that such background knowledge is not explicitly provided by the raw texts, and meanwhile, too abstract and complex to be extracted by the current language model techniques. In this case, images will help. For example, we could find many images where fire extinguishers appear in these scenes of public places. Therefore, this commonsense knowledge could be learned by perceiving the scene information of these images, and the corresponding question will be well answered. These analyses are in accordance with Minsky's statement in \citet{minsky2000commonsense}, `perhaps a good architecture theory based on multiple representations and multi-modal reasoning would help us to design better systems that allow us to study and understand commonsense reasoning.' 

Our approach, named Loire (\textbf{L}earning C\textbf{o}mmonsense from \textbf{I}mages for \textbf{Re}asoning), consists of two stages, i.e.~visual commonsense learning and knowledge-augmented reasoning. In the first stage, a scene layout generation task is conducted on a bi-modal data such as the representative benchmark COCO~\cite{lin2014microsoft}. Firstly, a text encoder Visual BERT (ViBERT for short) is employed to obtain the representation of a caption. ViBERT is then incorporated into the recurrent encoder-decoder structure for the labeled bounding box generation. This module is trained separately by a supervised learning approach, based on the ground-truth bounding boxes of images. In this way, the required visual commonsense knowledge will be encoded in ViBERT. In the following commonsense reasoning stage, the concerned text representations (such as question and answer in commonsenseQA) will be obtained by concatenating ViBERT and a traditional pre-trained language model, e.g. ~BERT. Then the language model is fine-tuned on the commonsense reasoning data, with ViBERT fixed as some prior knowledge. Experimental results on two commonsense reasoning tasks, i.e.~CommonsenseQA and WinoGrande \cite{sakaguchi2019winogrande}, demonstrate that the learnt commonsense from images brings improvements to traditional models, such as BERT fine-tune \cite{devlin2018bert} and RoBERTa fine-tune \cite{liu2019roberta}. We also give some case studies to show how the learned visual commonsense knowledge helps the reasoning process. 

To the best of our knowledge, we are the first to propose learning commonsense knowledge from images to facilitate the commonsense reasoning in NLP. The proposed model of using scene layout generation as the supervision demonstrates a preliminary exploration in this direction. Other methods like learning commonsense from retrieved relevant images could also be investigated. We believe this novel approach may provide a new perspective for commonsense reasoning in NLP.

\section{Related Work} \label{related-work}

\subsection{Commonsense reasoning Methods}
There are mainly two kinds of commonsense reasoning methods, knowledge base approach and raw text approach. 

Knowledge base approach makes use of the existing knowledge bases \cite{speer2017conceptnet, sap2019atomic} to conduct the commonsense reasoning process. Some methods regard knowledge base as a supplement and integrate extracted knowledge with information from the processed text.
For example, \citet{mihaylov2018knowledgeable} encodes external commonsense knowledge as a key-value memory. \citet{lv2019graph} and \citet{lin2019kagnet} extract knowledge from ConceptNet and Wikipedia to construct graphs, then use Graph Convolutional Network \cite{kipf2016semi} for modeling and inference. Other methods \cite{zhong2018improving, ma2019towards, ye2019align, li2019teaching} use knowledge bases as another corpus for pre-training, and then refining the models on task-specific contents. 

Besides extracting knowledge from knowledge bases, some other methods directly learn commonsense knowledge from raw texts. A common way is to use pre-trained language models. Recently, \citet{talmor2018commonsenseqa, da2019cracking, sakaguchi2019winogrande, zhou2019evaluating} have made comprehensive empirical studies and shown that pre-trained language models significantly outperform traditional methods on the task of commonsense reasoning. In addition, \citet{da2019cracking} proves that pre-trained language models have the ability to encode some commonsense knowledge in the embedding space through the attribute classification evaluation. However, they also show that the encoded commonsense knowledge is limited, which could be improved by introducing some supplementary data, like ConceptNet. 

Moreover, some methods propose to leverage additional text information/data for better commonsense reasoning. \citet{tandon2018reasoning} uses commonsense knowledge as constraints and large scale web corpus to steer the model away from unlikely predictions. 
\citet{rajani2019explain} incorporates the generated explanations into the training of language models for enhancement. \citet{xia2019incorporating} leverages two auxiliary relation-aware tasks to better model the interactions between question and candidate answers. \citet{chalier2020joint} proposes a multi-faceted model of commonsense knowledge statements to capture more expressive meta-properties. 

Different from the above approaches, we propose to learn commonsense from images and incorporate this visual knowledge into the following commonsense reasoning process.

\subsection{Bi-modal Language Models}
Recently, some transformer-based bi-modal language models~\cite{Su2019VL, li2019unicoder-vl:, alberti2019fusion, li2019visualbert, tan2019lxmert, lu2019vilbert} have been proposed to tackle with bi-modal reasoning problems in computer vision, such as visual question answering, visual commonsense reasoning, and image retrieval. They first encode image representation and text representation into a shared embedding space, then apply the joint embeddings for downstream reasoning. At first glance, these models are quite similar to ours. However, we should make it clear that they are totally different. The purpose of a bi-modal language model is to capture a cross-modal alignment between image and text to benefit the bi-modal task, which is only available when both image and text data are provided as input simultaneously. That is why they are usually popular in bi-modal scenarios like VQA. If we want to apply these models to commonsense reasoning in NLP, how to find corresponding images to the question, and how to employ the joint embeddings to the downstream NLP reasoning tasks is still unclear. Our model also adopts image data as a supplementary, but the modeling approach is different from bi-modal language models. We first encode the visual commonsense knowledge into ViBERT by the upstream layout generation process on a bi-modal data, then apply ViBERT as fixed prior knowledge to fine-tune the pre-trained language models for the downstream NLP reasoning tasks.

\section{Visual Commonsense Knowledge} \label{commonsense-in-image}

\begin{figure}
\centering
\includegraphics[width=0.31\columnwidth]{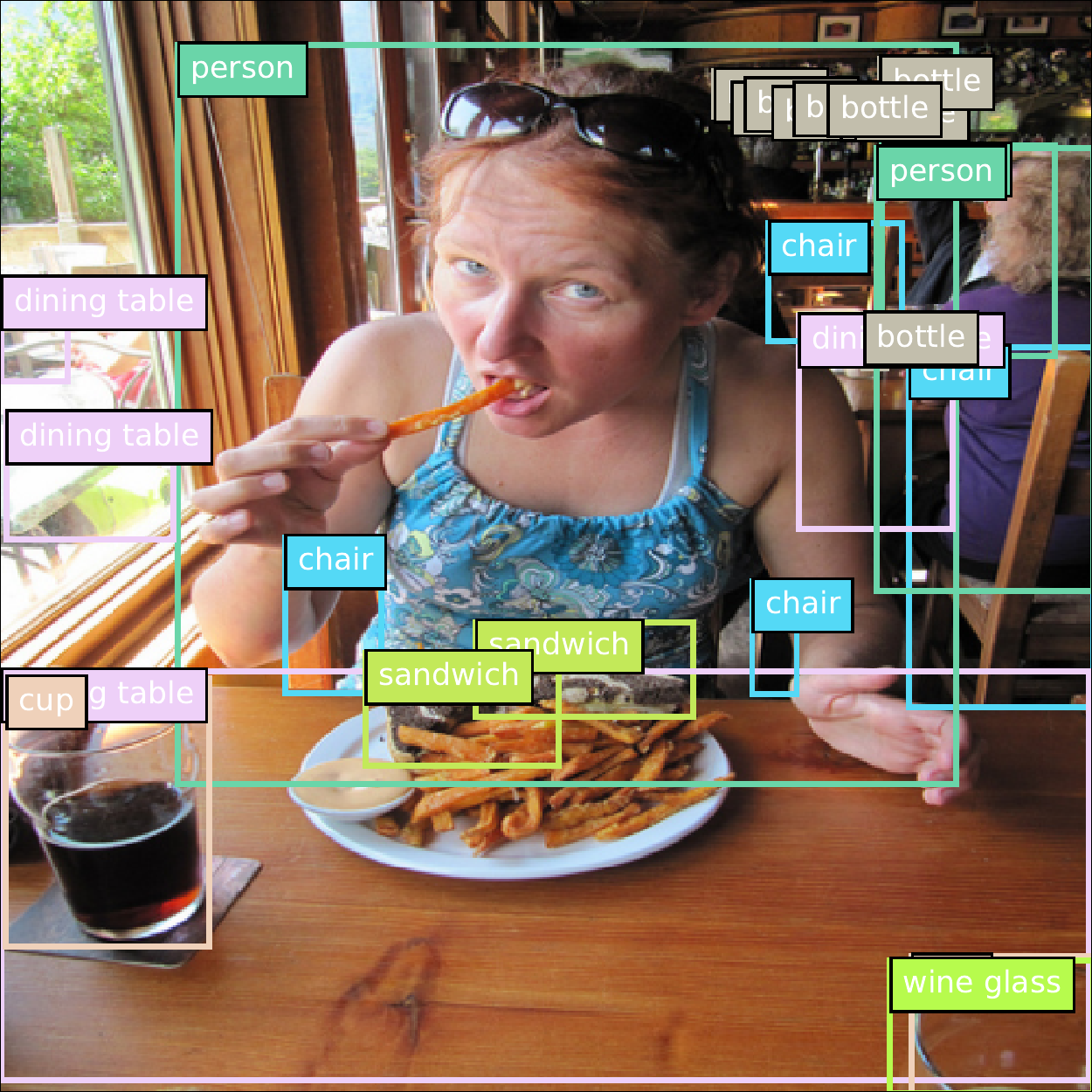}
\includegraphics[width=0.31\columnwidth]{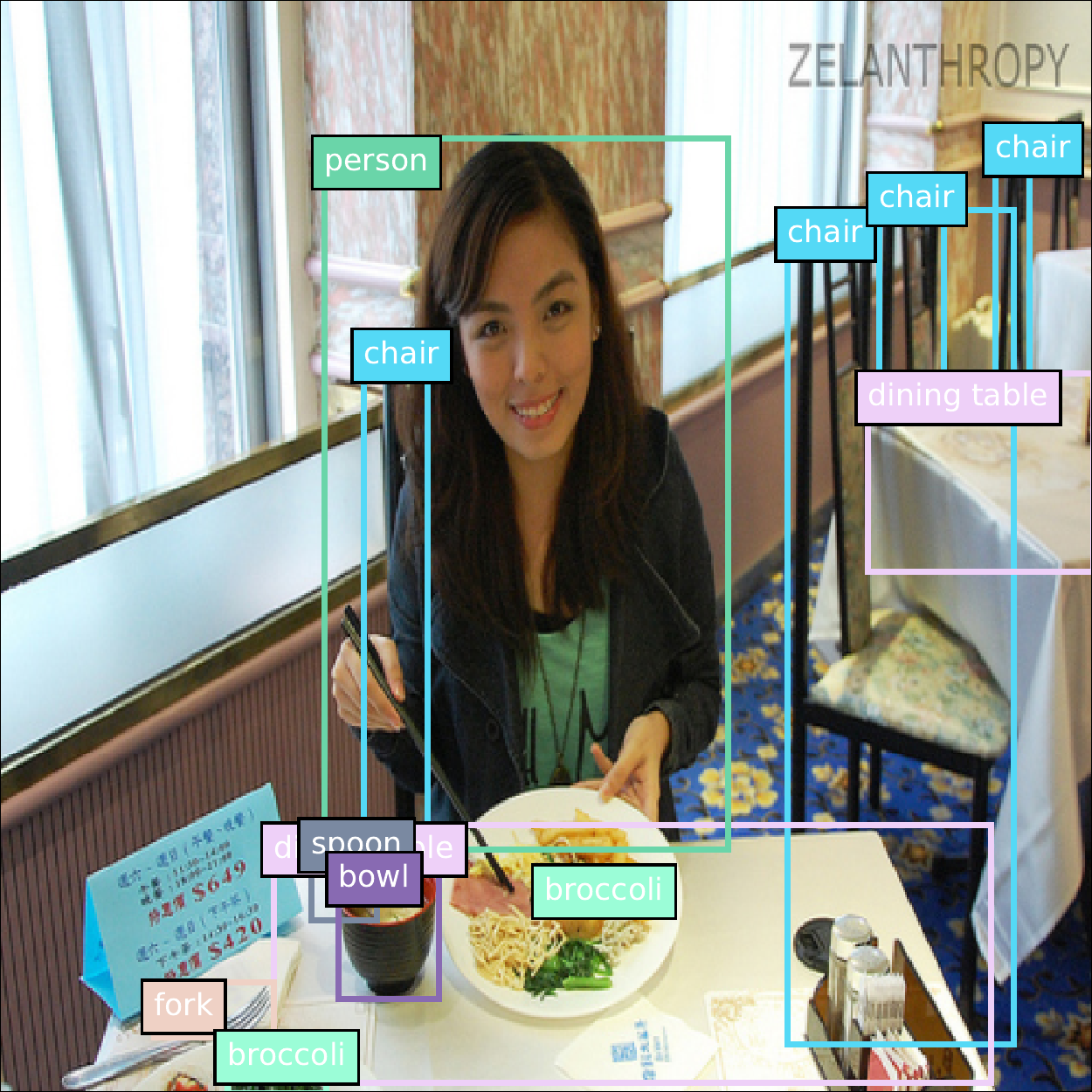}
\includegraphics[width=0.31\columnwidth]{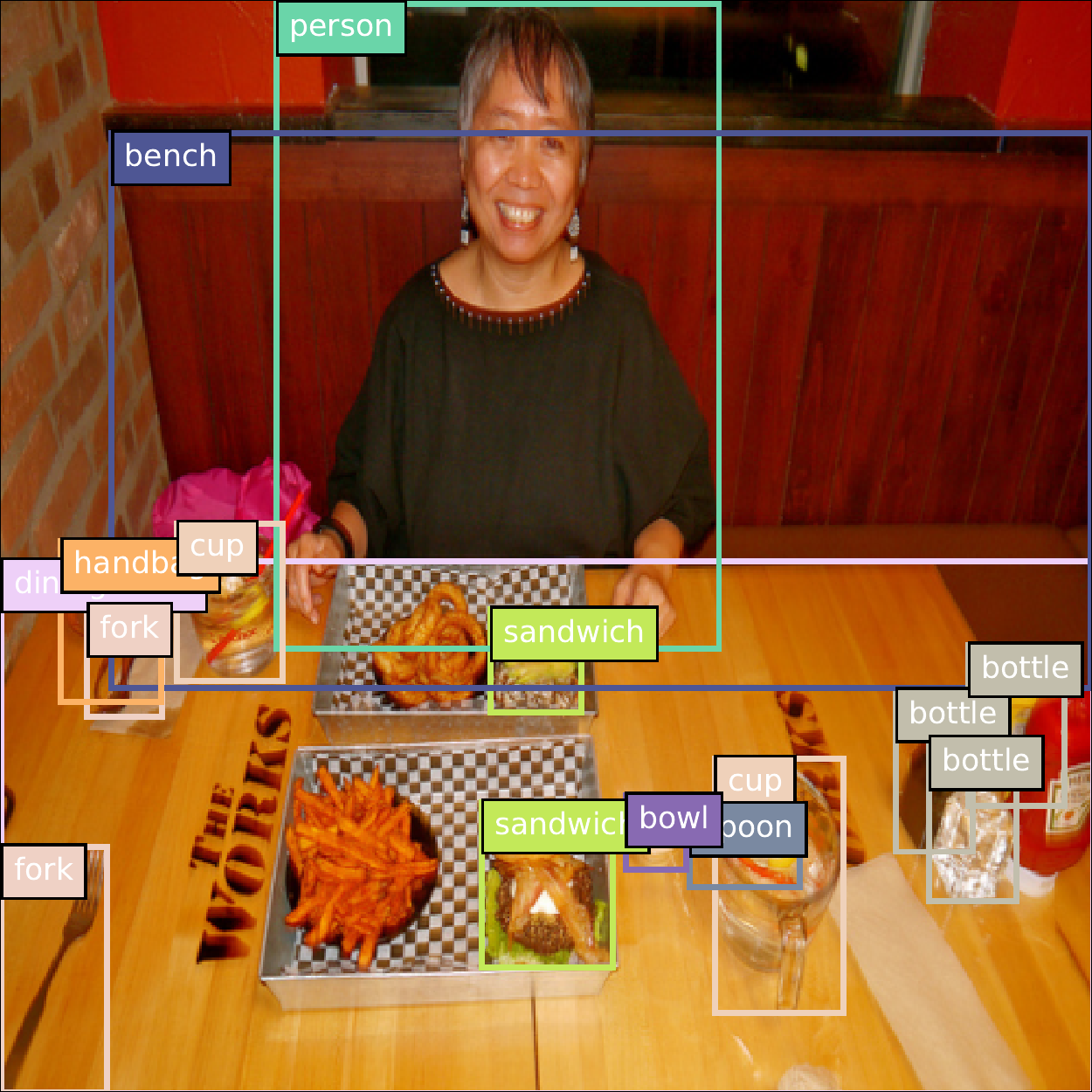}
\caption{Images and the associated bounding boxes from COCO with captions similar to `a woman eats in the restaurant'.}
\label{restaurant_example}
\vskip -0.25in
\end{figure}

Images are made up of individual pixels, which are detailed but sometimes noisy. Therefore, how to extract useful commonsense knowledge from images remains a challenging problem. Inspired by the knowledge base in NLP, where knowledge is usually represented as a triple to demonstrate the relation between two entities, we focus on the attributes and relations of the objects in images. Clearly, such information can be well captured by the scene layout. Take the sentence `a woman eats in the restaurant' as an example. Images related to this sentence are shown in the Figure~\ref{restaurant_example}. We can see that the scene layouts of these images, including bounding boxes and labels, contains a lot of useful information for commonsense reasoning:

(1) Size attributes and relations can be easily obtained by the bounding boxes in images. For instance, the bounding boxes of tableware, e.g. ~fork, cup, spoon are always smaller than the bounding boxes of the dining table.

(2) Position can be accurately captured by the coordinate of each bounding box, to help understand some abstract commonsense. For instance, through the relative positions between the bounding boxes of person and table, one can figure out what "next to" means. Besides, since the bounding boxes of person and table are always close in the eating scene, one can learn that if a person is eating, he will be next to the table instead of standing far away, which provides some detailed information for the abstract word `eating'.

(3) Co-occurrence relations of objects are expressed by the labels of bounding boxes. For instance, images of `a woman eats in the restaurant' often contain labels of table, chair, person, food and tableware. So from the co-occurrence of these objects, one can infer that it is in a dinner or restaurant scenario, which offers rich context information for the abstract word `eating'.

From the above analysis, images usually contain rich scene information, such as size, position and co-occurrence relations, which are useful for understanding the commonsense knowledge hidden in language. So we propose to learn such visual commonsense knowledge and incorporate them into the commonsense reasoning models in NLP.

\section{Our Approach: Loire}
Now we introduce Loire, which includes two stages, i.e.~visual commonsense learning and knowledge-augmented reasoning. 

\begin{figure}
\begin{center}
  \centerline{\includegraphics[width=0.9\columnwidth]{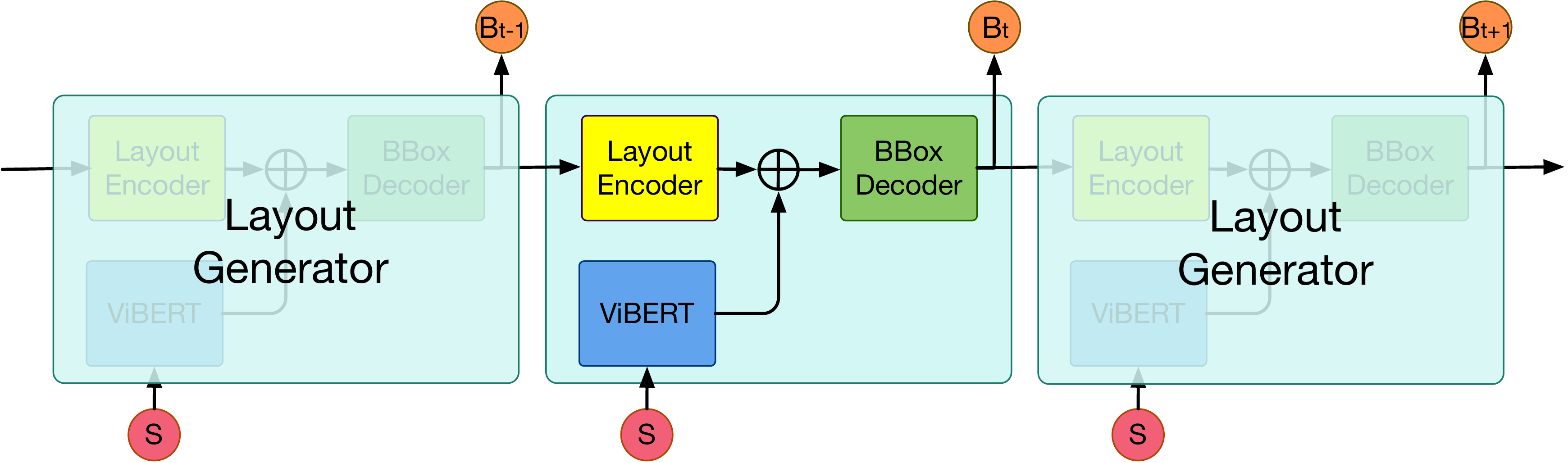}}
  \caption{The recurrent structure of the visual commonsense learning stage.}
\label{overall-for-layout}
\end{center}
\vskip -0.4in
\end{figure}

\subsection{Visual Commonsense Learning} \label{scene-layout-generation}
\begin{figure}
\begin{center}
  \centerline{\includegraphics[width=0.9\columnwidth]{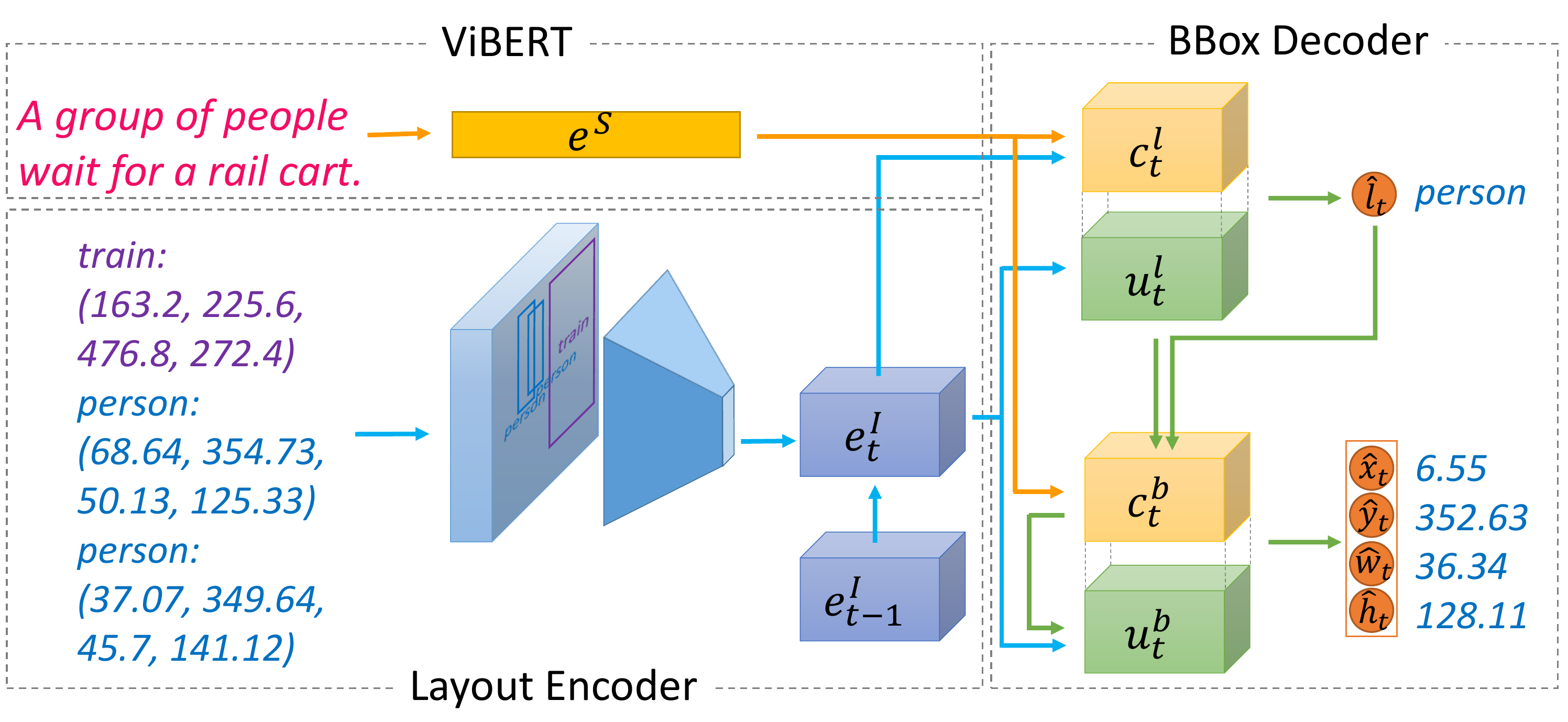}}
  \caption{An illustration of the $t$-th layout generation.}
\label{onestep-for-layout}
\end{center}
\vskip -0.4in
\end{figure}

The visual commonsense learning stage is conducted on bi-modal data, like the typical image caption data COCO. For a given image, the required scene layout is generated by a sequence-to-sequence approach, shown in Figure~\ref{overall-for-layout} and ~\ref{onestep-for-layout}. This module consists of a text encoder, namely ViBERT, to map the input
sentence to a latent representation, a layout encoder to encode the current generated scene layout, and a bounding box decoder to generate the next bounding box and its label. 

Specifically, we make the following notations. Let the input image caption be $S = \{w_1, w_2, \dots, w_L\}$, where $w_i$ stands for the $i$-th word in the sentence, and $L$ is the sentence length. The output is a set of labeled bounding boxes $B_{1:T} = \{B_1, ..., B_T \}$, with each labeled bounding box $B_t$ contains the position, size and category label of a corresponding object at the $t$-th step. So we denote $B_t = (b_t,l_t)$, where $b_t = [b_t^x, b_t^y, b_t^w, b_t^h] \in \mathbb{R}^4$ stands for 2-dimensional coordinates, width and height, respectively. $l_t \in \{0,1\}^{C+1}$ is a one-hot vector to indicate the category label for an object, and the additional $C+1$ class is defined as a special indicator for the end of generation.

\subsubsection{ViBERT: Text Encoder}
The text encoder ViBERT is fine-tuned from BERT, which is a popular pre-trained language model introduced in \citet{devlin2018bert}. The network structure is a typical transformer-based architecture containing multiple transformer blocks of multi-headed scaled dot product attention and fully connected layers~\cite{vaswani2017attention}. It has been proven to be effective in many natural language processing tasks.

To adapt to the setting of BERT, the image caption is preprocessed as follows. The special tokens `[CLS]' and `[SEP]' are inserted into the beginning and the end of the sentence, to obtain $S = \{w_0, w_1, ..., w_{L+1}\}$, where $w_0, w_{L+1}$ stands for $\textrm{[CLS]}$ and $\textrm{[SEP]}$, respectively. After that, each word $w_i$ is mapped to its word embedding vector $e_i^S$ by ViBERT, so that $e(S) = \{e_0^S,e_1^S,...,e_{L+1}^S\}$.
With BERT, the output of `[CLS]' from the last transformer layer is fed into a pooler layer to obtain the representation of the whole sentence $e^S$,
\begin{equation}
    e^S = \textrm{tanh}(f(e_0^S)),
\end{equation}
where $f$ is a single-layer perceptron.

\subsubsection{Layout Encoder}
At each time step $t$, a layout encoder is utilized to encode the state of the current generated layout $B_{0:t-1}$. Specifically, we construct a layout matrix $I_{t-1} \in \{0,1\}^{C \times W \times H}$, where $W, H$ are width and height of this layout respectively. The value of $i_{lwh}$ in $I_{t-1}$ indicates whether the bounding box of object $l$ covers the pixel at coordinate $[w, h]$. A blank layout without any object is used to initialize $B_0$. A layout encoder takes layout matrix and previous layout representation as inputs, and uses a convolutional GRU architecture to output the representation of the current layout $e^I_t$ as follows:
\begin{equation}
  e^I_t = \textrm{ConvGRU}(I_{t-1}, e^I_{t-1}).
\end{equation}

\subsubsection{Bounding Box Decoder}
At each time step $t$, a bounding box decoder is used to predict the labeled bounding box of next object, based on the caption representation $e^S$ from ViBERT and the current layout representation $e_t^I$ from the layout encoder. Specifically, we decompose the conditional joint bounding box probability as $p(b_t,l_t|s, B_{0:t-1}) = p(l_t|S, B_{0:t-1}) p(b_t|S, B_{0:t-1},l_t)$. The decoder firstly samples a class label $l_t$ according to $p(l_t|S, B_{0:t-1})$:
\[
  p(l_t|s,B_{0:t-1}) = \textrm{Softmax}(g(u_t, c_t)),
\]
\[
  u_t^l = \phi^l(e^I_t, e^S),\;\;
  c_t^l = \varphi^l([u_t^l; l_{1:t-1}], e(S)), 
\]
where $g$ is a two-layer perceptron, $\phi^l$ is a Convolution network~\cite{xu2015show} with spatial attention on $e^I_t$, and $\varphi^l$ is a text-based attention module~\cite{luong2015effective}, which is used to focus on different parts of the caption. 

After that, the decoder tries to find out $b_t$ for object $l_t$ based on $p(b_t|S,B_{0:t-1},l_t)$, which is obtained by a regression network $\theta$ with $\hat{b}_t = (\hat{x}_t, \hat{y}_t, \hat{w}_t, \hat{h}_t) = \theta(c_t^b, u_t^b)$. The parameters $c_t^b$ and $u_t^b$ are represented similarly to $u_t$ and $c_t$. That is,
\[
   u_t^b = \phi^b (e^I_t, c_t^b), \;\;c_t^b = \varphi^b([u_t^l; l_t], e(S)),
\]
where $\phi^b$ is an image-based attention module to find an appropriate position, and $\varphi^b$ is another text-based attention module but focuses more on the contents related to the current object.

\subsubsection{Training}
To reduce the expensive training ViBERT from scratch, we initialize ViBERT with the parameter weights of BERT$_{BASE}$ released by Google~\footnote{\url{https://github.com/google-research/bert}}. Then the scene layout generator can be trained by minimizing the negative log-likelihood of the ground-truth object labels and the mean-square error of the ground-truth bounding box coordinates as follows:
\begin{equation}
  \mathcal{L}_{layout} = \sum_t\big(||\hat{b_t}-b_t^*||_2 - \log p(l_t^*) \big),
\end{equation}
where $b_t^*$ and $l_t^*$ stands for the ground-truth bounding box and label, respectively. As for the generation order, we have observed that the model is difficult to converge with unfixed order, which may be caused by some dependencies among different bounding boxes. So we follow the existing image generation methods and simply fix the order from bottom to top, left to right. 

It should be noted that although we use BERT as a text encoder on image captions, we do not optimize the objective of the language model, i.e. ~ the masked language model (MLM) objective. This is to avoid the possibility that the improvement of downstream reasoning task is due to the use of more text data, instead of visual commonsense knowledge from images. In our experiments, we have conducted some ablation studies to validate this point.

\subsection{Knowledge-Augmented Reasoning} \label{knowledge-argumented-reasoning}
After using scene layout generation to encode visual commonsense knowledge into ViBERT, we can apply ViBERT as a fixed prior to enhance the downstream commonsense reasoning tasks. 

Here we use commonsenseQA as an example to demonstrate our method. For a given question $q_i \in Q$, where $Q$ is the question set, and its candidate answers $A_i = \{a_i^1, \dots, a_i^n\}$, where $n$ denotes the number of choices, a common existing method is to first concatenate question and each candidate answer to a raw representation $[q_i;a_i^j]$. Then a pre-trained language model is applied to obtain a semantic representation, denoted as $E_{i,j}^{(1)} = \textrm{LM}([q_i;a_i^j])$. In our method, ViBERT is applied on the raw representation $[q_i;a_i^j]$ to obtain a image scene-aware text representation, denoted as $E_{i,j}^{(2)} = \textrm{ViBERT}([q_i;a_i^j])$. Since the two representations are not always in the same space, we use a projection matrix $M$ to project $E_{i,j}^{(2)}$ to the space of $E_{i,j}^{(1)}$. After that, they are simply concatenated and fed into a linear layer to compute the probability $p(a_i^j|p_i)$ as follows.
\[
  \textrm{score}(a_i^j) = h(E_{i,j}^{(1)}; M^T E_{i,j}^{(2)}]),
\]
\[
  p(a_i^j|p_i) = \textrm{Softmax}(\{\textrm{score}(a_i^j)\}_j),
\]
where $h$ is a simple linear layer for classification, and the parameters of both language model and the linear layer will be fine-tuned on the downstream commonsense reasoning task. Specifically, in the training process, the objective is to minimize the negative log-likelihood of ground-truth answers $a_i^*$ as follows. After that, the choice with the highest score will be selected as the answer. 
\begin{equation}
  \mathcal{L}_{qa} = -\sum_i \log p(a_i^*|q_i).
\end{equation}

\section{Experiments} \label{experiment}
This section demonstrates our experiments on two commonsense reasoning tasks, i.e.~comonsense question answering and pronoun resolution.

\subsection{Datasets}

\textbf{CommonsenseQA}\footnote{\url{https://www.tau-nlp.org/commonsenseqa}}~\cite{talmor2018commonsenseqa} is a typical commonsense question answering dataset, which consists of 12,102 natural language questions generated from ConceptNet. It covers various types of commonsense knowledge, including spatial, causal, social, and activity, etc. Each question has five candidate answers. Table~\ref{csqa-examples} shows 3 question-answering examples. In our experiments on this dataset, we use the official random-split setting for fair comparisons with the reported results on CommonsenseQA’s leaderboard. 

\textbf{WinoGrande}\footnote{\url{https://leaderboard.allenai.org/winogrande/submissions/get-started}}~\cite{sakaguchi2019winogrande} is a challenging pronoun resolution dataset extended from the original Winograd Schema Challenge \cite{levesque2012winograd}. The task is about resolving a pronoun (represented as a blank line) to one of its two probable co-referents in the sentence. For this task, each sentence is treated as a fill-in-the-blank question with binary choices. The line in the sentence is replaced by each option, and the model is required to provide the likelihood for the two resulting sentences for determination. In the training set of WinoGrande, there are five different sizes, i.e.~XS(160), S (640), M (2,558), L (10,234) and XL (40,398). We experiment on all the five sizes and report their results for analysis.

\subsection{Experimental Settings}
For the upstream scene layout generation module, we train our \textbf{ViBERT} on 2 Nvidia K80 GPUs with a batch size of 32 for 15 epochs. The learning rate is $5e^{-5}$, and the optimizer is Adam with StepLR schedule, where the step size is 3 and $\gamma$ is 0.8. In the training process, the bi-modal data COCO~\cite{lin2014microsoft} is used to train our layout generation model. COCO consists of 123,287 images over 80 object categories, and each image is associated with instance-wise annotations and 5 image captions. For better training, we ignore small objects and filter images with more than 20 objects. This leaves us 119,146 images. We use the official train and validation splits, and set a max sequence length as 128.

For the downstream commonsense reasoning module, we choose BERT and RoBERTa as our baseline models, which are the fundamental and competitive models for NLP tasks. 

\textbf{BERT}~\cite{talmor2018commonsenseqa} is a powerful contextualized word representation model and has been proven helpful in many NLP tasks. We apply uncased BERT$_{\textrm{BASE}}$ to downstream commonsense reasoning tasks by encoding each question and its candidate answers as a series of delimiter-separated sequences, i.e. ~`[CLS] question [SEP] choice [SEP]' for CommonsenseQA and `[CLS] segment1 [SEP] option segment2 [SEP]' for WinoGrande. Then the representation of `[CLS]' is then fed into a BERT-Pooler and converted to predictions by a linear classification layer.

\textbf{RoBERTa}~\cite{liu2019roberta} is similar to BERT, but is usually pre-trained with a larger amount of training data and different techniques such as dynamic masking. Besides RoBERTa$_{\textrm{BASE}}$, we also compare with a fine-tuned RoBERTa$_{\textrm{LARGE}}$ following the implementation released in fairseq~\footnote{\url{https://github.com/pytorch/fairseq/tree/master/examples/roberta/commonsense_qa}}. And according to fairseq, we prepend a prefix of Q: to the question and A: to the answer for CommonsenseQA, which was found to be helpful.

\textbf{Loire} By using BERT and RoBERTa as a language model for text, we concatenate the representations from ViBERT and the pre-trained language model, and obtain two versions of our model, denote as \textbf{Loire-BERT} and \textbf{Loire-RoBERTa}, respectively. Since ViBERT is a static feature extractor and doesn't need to be fine-tuned in the downstream reasoning tasks, our running time is similar to the baselines without extra time cost.

We train all models on 2 Nvidia K80 GPUs using AdamW~\cite{loshchilov2018fixing} with WarmupLinearSchedule approach~\cite{he2016deep} for optimization, where the warmup percentage is set to 0.1 and 0.05 for BERT and RoBERTa, respectively.
We use grid-search for hyper-parameters tuning. The learning rate, number of epochs and batch-size are chosen from $\{1, 2\} \times e^{-5}$, $\{3, 5, 8\}$, and $\{8, 16, 32\}$. 
The best development set accuracy from 5 random restarts of fine-tuning is reported, with the standard deviation. The best models on the development dataset are then submitted to the official private test dataset to return the test results. All our code and data are publicly available at \url{https://github.com/VickiCui/Loire}.

\subsection{Experimental Results}
On the dev set, the accuracy of the layout generation for label prediction is 63.4\%, and the mean square error for bbox prediction is 0.015 (the coordinates of bbox have been standardized between 0 and 1). This shows that the layout generator has a good performance and can generate good quality scene layouts, and the model does learn the corresponding knowledge.

\begin{table}
\caption{Results on CommonsenseQA (\%), where `*' indicates the reported result from the leaderboard.}
\label{csqa_result}
\begin{center}
\begin{small}
\begin{adjustbox}{max width=1.\linewidth}
\begin{tabular}{lccc}
    \toprule
    Model &  Dev Acc. & Dev Avg.& Test Acc.\\
    \midrule
    \citet{ott2019fairseq}  & - & - & \textbf{72.1$^*$} \\
    \midrule
    RoBERTa$_{\textrm{LARGE}}$  & 77.47 & 76.65$\pm{0.58}$ & 71.58 \\
    Loire-RoBERTa$_{\textrm{LARGE}}$  & \textbf{77.94} & \textbf{77.56$\pm{0.28}$} & \textbf{71.93} \\
    
    RoBERTa$_{\textrm{BASE}}$  & 65.47 & 64.96$\pm{0.62}$ & 59.82 \\
    Loire-RoBERTa$_{\textrm{BASE}}$   & 66.67 & 66.12$\pm{0.47}$ &60.61 \\
    
    BERT$_{\textrm{BASE}}$  & 59.71 & 58.95$\pm{0.65}$ & 53.00$^*$ \\
    Loire-BERT$_{\textrm{BASE}}$  & 61.19 & 60.07$\pm{0.58}$ & 54.91  \\
    \midrule
    Human & - & - & 88.00 \\
    \bottomrule
\end{tabular}
\end{adjustbox}
\end{small}
\end{center}
\vskip -0.15in
\end{table}

\begin{table}
\caption{Results on WinoGrande with 5 training sizes, where `*' indicates the reported result from the leaderboard.}
\label{wsc_result}
\vskip -0.25in
\begin{center}
\begin{small}
\begin{adjustbox}{max width=1.\linewidth}
\begin{tabular}{lccccc}
    \toprule
    Model & XS & S & M & L & XL\\
    
    \midrule
     & \multicolumn{5}{c}{Dev Acc. (\%)} \\
    BERT$_{\textrm{BASE}}$ & 50.76 & 51.61 & 52.81 & 55.26 & 60.19   \\
    Loire-BERT$_{\textrm{BASE}}$ & \textbf{51.61} & \textbf{52.34} & \textbf{53.9} & \textbf{56.74} & \textbf{61.50}   \\
    RoBERTa$_{\textrm{BASE}}$ & 51.72 & 54.71 & 57.91 & 62.52 & 67.94   \\
    Loire-RoBERTa$_{\textrm{BASE}}$ & \textbf{53.26} & \textbf{55.18} &\textbf{ 58.93} & \textbf{64.09} & \textbf{69.21 }  \\
    RoBERTa$_{\textrm{LARGE}}$ &  52.40 & 61.95 & 68.67 & 75.14 & 79.08   \\
    Loire-RoBERTa$_{\textrm{LARGE}}$ &  \textbf{52.64} & \textbf{63.06} & \textbf{70.40} & \textbf{76.56} & \textbf{81.06}   \\
    
    \midrule
     & \multicolumn{5}{c}{Test Acc. (\%)} \\
    BERT$_{\textrm{BASE}}$ & 49.75 & \textbf{49.75} & 49.01 & 51.50 & 54.73   \\
    Loire-BERT$_{\textrm{BASE}}$ & \textbf{49.86} & 49.29 & \textbf{52.07} & \textbf{53.88} & \textbf{59.54}   \\
    RoBERTa$_{\textrm{BASE}}$ & 50.93 & 52.01 & \textbf{57.67} & 61.35 & 65.42 \\
    Loire-RoBERTa$_{\textrm{BASE}}$ & \textbf{53.42} & \textbf{53.42} & 56.82 & \textbf{62.31} & \textbf{67.12}\\
    \citet{levesque2012winograd} &  50.37 & 58.63 & 67.57 & 74.70 & \textbf{79.12}   \\
    \citet{yang2020g} &  \textbf{55.04} & 62.37 & 66.72 & 74.19 & 78.21   \\
    Loire-RoBERTa$_{\textrm{LARGE}}$ &  53.14 & \textbf{63.27} & \textbf{70.51} & \textbf{76.12} & 77.99   \\
    \bottomrule
\end{tabular}
\end{adjustbox}
\end{small}
\end{center}
\vskip -0.25in
\end{table}

Table~\ref{csqa_result} shows the experimental results on CommonsenseQA. From the results, we can see that our approach leads to a 1.91\%, 0.79\% and 0.35\% improvement in terms of accuracy on the test set, as compared with BERT$_{\textrm{BASE}}$, RoBERTa$_{\textrm{BASE}}$ and RoBERTa$_{\textrm{LARGE}}$ respectively. Similar results have been observed on the development set. Besides, the standard deviation of several random results on the development set becomes smaller when using Loire, which demonstrates better stability. Someone may argue that the improvement is marginal as compared with RoBERTa$_{\textrm{LARGE}}$, and our result is worse than the best result of RoBERTa$_{\textrm{LARGE}}$ on the leaderboard \cite{ott2019fairseq}. It should be noted that the best result of RoBERTa$_{\textrm{LARGE}}$ on the leaderboard is based on validation performance after 100 trials. However, we only conducted five trials in our experiments due to our limited computing resources. The purpose of this paper is to propose a new perspective of learning commonsense from the image, rather than achieving a SOTA result. We can clearly see some improvement from the comparison with the baseline models. It is acceptable that when using more complicated language models, the effect of visual knowledge will be weakened. However, there are indeed some methods to improve the current results, which will be investigated in our future work. For example, we have filtered out small objects to make the training easier, which may result in insufficient details. Besides, the adopted bi-modal data COCO is very limited, with only 80 categories of objects. On the one hand, the coverage of the commonsense may be restricted. On the other hand, the layouts generated by our model may not be very accurate for some objects. For instance, the generated layout of `laundry' is `a suitcase' since COCO does not contain clothes in our case study. We plan to employ larger data such as Visual Genome~\cite{krishna2017visual} to tackle this problem.

Table~\ref{wsc_result} shows the experimental results on WinoGrande. Specifically, five models are trained on five different training data sizes separately, and the development set and test set are identical for all models. As for the accuracy of the development set, We can see that Loire achieves consistent performance improvements across different sizes of training data, as compared with both BERT$_{\textrm{BASE}}$, RoBERTa$_{\textrm{BASE}}$ and RoBERTa$_{\textrm{LARGE}}$. While for the test accuracy (\citet{levesque2012winograd} and \citet{yang2020g} are two test results of RoBERTa$_{\textrm{LARGE}}$ from the leaderboard), except for a few ones, Loire consistently outperforms the corresponding baselines on across different sizes of training data. These results show the effectiveness of incorporating visual scene knowledge for commonsense reasoning.

\subsection{Ablation Study}

\begin{table}
\caption{Accuracy (\%) of different models on CommonsenseQA development set.}
\label{ablation_study}
\begin{center}
\begin{small}
\begin{tabular}{lccc}
    \toprule
      Model. & Dev Acc. & Dev Avg\\
    \midrule
    BERT$_{\textrm{BASE}}$  & 59.71 & 58.95$\pm{1.03}$\\
    \hspace{0.9em}+BERT$_{\textrm{BASE}}^*$  & 59.89 & 59.12$\pm{0.65}$ \\
    \hspace{0.9em}+BERT$_{\textrm{CAPTION}}$  & 60.29 & 59.47$\pm{0.60}$\\
    \hspace{0.9em}+ViBERT (ours)  & \textbf{61.19} & \textbf{60.07$\pm{0.58}$} \\
    \bottomrule
\end{tabular}
\end{small}
\end{center}
\vskip -0.25in
\end{table}

In order to validate that the performance improvement owes to the introduction of learned visual commonsense knowledge, rather than using more parameters or data, we conduct the following ablation studies on CommonsenseQA. The results are shown in Table~\ref{ablation_study}, where `+ViBERT' denotes Loire.

Firstly, we study whether the improvement owes to the use of additional parameters. To this end, we compare with the BERT$_{\textrm{BASE}}$ concatenated with freeze BERT$_{\textrm{BASE}}^*$ features, in which the parameters are set to be the same as BERT$_{\textrm{BASE}}$+ViBERT. From the results, we can see that, under the same setting, the accuracy of BERT$_{\textrm{BASE}}$ concatenated with freeze BERT$_{\textrm{BASE}}^*$ features is $59.89\%$ on dev set, which is worse than ours. 

Then we study whether the improvement owes to the use of additional text data, i.e.~captions in COCO. We first fine-tune a BERT$_{\textrm{BASE}}$ model on COCO captions with MLM objective, denoted as BERT$_{\textrm{CAPTION}}$. Then we concatenate it with BERT$_{\textrm{BASE}}$, to perform a similar downstream fine-tuning as in Loire-BERT$_{\textrm{BASE}}$. We also randomly initialized the model 5 times. The best dev result is $60.29\%$, which is worse than Loire.

In summary, these ablation studies prove that the commonsense knowledge learned form images, rather than the introduction of more parameters or text data, is responsible for the improvements.

\subsection{Case Study}

\begin{table}[t]
\caption{Case study examples from the dev set of CommonsenseQA.}
\label{csqa-examples}
\begin{center}
\begin{small}
\begin{tabular}{l p{5cm}}
\toprule
Question1: & The man got a pail to catch the draining motor oil, where was he likely doing this at home?\\
Choices1: & \textbf{(A)~garage} (B)~hardware store (C)~utility room (D)~wishing well (E)~laundry\\
\midrule
Question2: & Where would a person be doing when having to wait their turn?\\
Choices2: & (A)~have patience (B)~get in line (C)~sing \textbf{(D)~stand in line} (E)~turn left\\
\midrule
Question3: & Where would you find magazines along side many other printed works?\\
Choices3: & (A)~doctor \textbf{(B)~bookstore} (C)~market (D)~train station (E)~mortuary\\
\bottomrule
\end{tabular}
\end{small}
\end{center}
\vskip -0.25in
\end{table}

To understand what type of commonsense knowledge is learned by Loire, we analyze the relations between question concept and answer concept in CommonsenseQA according to ConceptNet. For the part of the questions that are done right by our model but wrong by the text-only model, which can be seen benefit from images, the top three relation types are AtLocation (36.4\%), Causes (12.7\%) and RelatedTo (8.5\%). These relationships can indeed be expressed through the scenes shown in the images. So this is accordant with our motivation, and the introduction of images can indeed play a complementary role. For complete statistics of relation types, please see Appendix A.

Table~\ref{csqa-examples} gives three examples in the development set of CommonsenseQA that benefit from visual commonsense knowledge. To better understand how visual commonsense helps, we generate the layout for each pair of question and choice by the trained upstream layout generator. Figure~\ref{case-study-layout} shows the layouts of Question1 and its choices, and others can be found in Appendix B due to space limitations.

Take the first question as an example, language models mainly rely on word co-occurrence or semantics for modeling, so they are easy to wrongly choose 'utility room' as the answer. That is because it is difficult to capture the commonsense of `got a pail to catch the draining motor oil in garage' from the pure language. From Figure~\ref{case-study-layout}, we can see that the layout of question, the correct answer `garage' and the wrong answer `utility room' are 'a person' with `a truck', `cars', and 'chairs' and `old televisions', respectively. That is to say, we can learn from images that `got a pail to catch the draining motor oil' usually happen with the scene that a person is with a truck. By encoding this knowledge into ViBERT, it is easy for the language model to connect the similarity between `truck' and `cars', so Loire is able to choose the correct answer 'garage', instead of 'utility room'.

\begin{figure}
\begin{center}
  \centerline{\includegraphics[width=0.9\columnwidth]{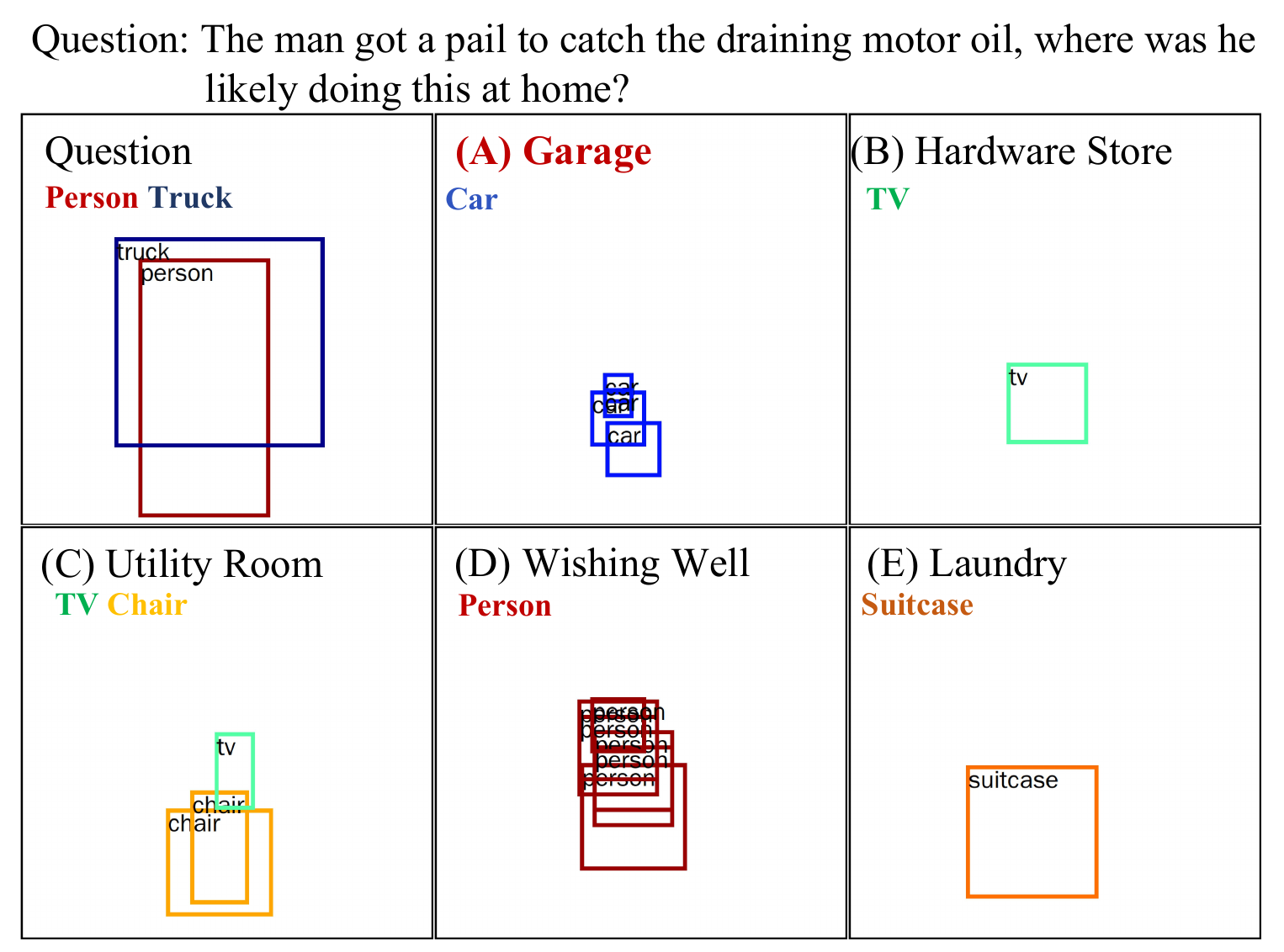}}
  \caption{Scene layout of the first example in Table~\ref{csqa-examples}.}
\label{case-study-layout}
\end{center}
\vskip -0.4in
\end{figure}

\section{Conclusion} \label{conclusion}
In this paper, we propose a novel two-stage pipeline approach Loire to learn commonsense from images. In the first stage, a text representation model ViBERT is trained in the bi-modal sequence-to-sequence approach for scene layout generation on COCO. Therefore, visual commonsense knowledge like spatial relations will be encoded in ViBERT by the supervision of caption and image layout. After that, ViBERT is concatenated with a pre-trained language model to perform a knowledge-augmented reasoning process. Experimental results show that Loire outperforms the current state-of-the-art language models BERT and RoBERTa on two NLP commonsense reasoning tasks, i.e.~commonsense question answering data CommonsenseQA and pronoun resolution data WinoGrande. The ablation and case study further show that the improvements are truly owing to the learned visual commonsense knowledge, and how this knowledge helps the NLP reasoning process.

The current approach is a preliminary study on the proposed direction of using images to automatically learn commonsense knowledge to facilitate the NLP reasoning tasks, which could be modified from the following aspects to further improve the empirical performances. Firstly, larger bi-modal data could be employed to learn more commonsense required in the reasoning task. Secondly, other bi-modal methods instead of training ViBERT by the supervision of scene layout generation may be investigated. Thirdly, how to design intrinsic evaluation to help to understand what is learned by Lorie is still challenging and will be considered in the future.

\section*{Acknowledgement}
This work was supported by the National Key R\&D Program of China (2020AAA0105200), the National Natural Science Foundation of China (NSFC) under Grants No. 61722211, 61773362, 61872338, and 61906180, the Lenovo-CAS Joint Lab Youth Scientist Project, and the Foundation and Frontier Research Key Program of Chongqing Science and Technology Commission (No. cstc2017jcyjBX0059), the Tencent AI Lab Rhino-Bird Focused Research Program (No. JR202033).

\bibliography{emnlp2020}
\bibliographystyle{acl_natbib}

\onecolumn
\appendix

\section{Relation Types Analysis}
\begin{table}[h]
\caption{The relation types that benefit from images.}
\label{rel-types}
\begin{center}
\begin{small}
\begin{tabular}{lc|lc|lc}
    \toprule
      Relations & Proportion(\%) & Relations & Proportion(\%) & Relations & Proportion(\%)\\
    \midrule
    MotivatedByGoal & 1.7 & HasProperty & 0.8 & CausesDesire  & 4.2 \\
    HasPrerequisite & 5.1 & Desires & 0.8 & CapableOf  & 5.9 \\
    HasSubevent & 5.1 & RelatedTo & 8.5 & DistinctFrom  & 1.7 \\
    HasA & 1.7 & NotDesires & 0.8 & HasLastSubevent  & 0.8 \\
    PartOf & 2.5 & UsedFor & 4.2 & AtLocation  & 36.4 \\
    FormOf & 0.8 & Antonym & 5.9 & Causes  & 12.7 \\
    \bottomrule
\end{tabular}
\end{small}
\end{center}
\end{table}

\section{Layout Examples}

\begin{figure}[h]
\begin{center}
  \centerline{\includegraphics[width=0.9\columnwidth]{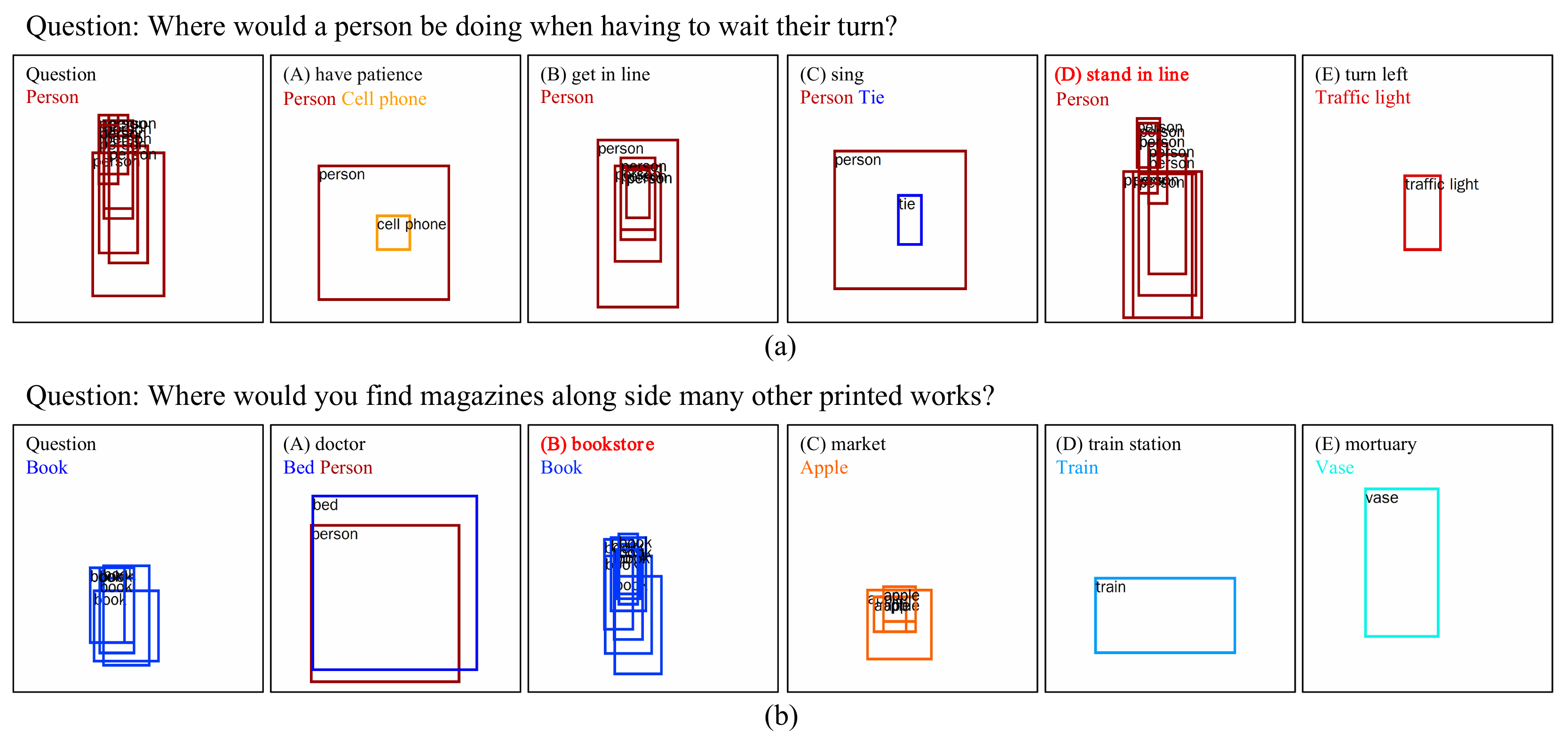}}
  \caption{Layout example that generated by scene layout generator. Images in the first column are layouts for questions. The layout for each choice is given in the other images.}
\label{case-study-layout}
\end{center}
\vskip -0.2in
\end{figure}

In this appendix, we visualize two more layout examples to show how the learned visual commonsense knowledge in our model helps the commonsense reasoning process.

As shown in Figure~\ref{case-study-layout} (a), according to the question, we can get a layout "a line of people", which is similar to the layouts of correct answer `stand in line' and choice `get in line'. In this case, visual commonsense knowledge helps the model eliminate irrelevant choices. 

As shown in Figure~\ref{case-study-layout} (b), we obtain the layout `a row of books' for the question, which exactly matches the layout of the answer `bookstore'. In this case, the visual commonsense knowledge directly helps the model get the correct answer.
\end{document}